\documentclass[conference]{IEEEtran}
\IEEEoverridecommandlockouts
\usepackage{cite}
\usepackage{amsmath,amssymb,amsfonts}
\usepackage{algorithmic}
\usepackage{graphicx}
\usepackage{textcomp}
\usepackage{xcolor}
\def\BibTeX{{\rm B\kern-.05em{\sc i\kern-.025em b}\kern-.08em
    T\kern-.1667em\lower.7ex\hbox{E}\kern-.125emX}}
\begin{document}

\title{A Comparative Study on COVID-19 Fake News Detection Using Different Transformer Based Models
}

\author{
\IEEEauthorblockN{Sajib Kumar Saha Joy}
\IEEEauthorblockA{\textit{Department of Computer Science and Engineering} \\
\textit{Ahsanullah University of Science and Technology}\\
Dhaka, Bangladesh \\
joyjft@gmail.com} \\

\IEEEauthorblockN{Dibyo Fabian Dofadar, Riyo Hayat Khan, Md. Sabbir Ahmed, Rafeed Rahman}
\IEEEauthorblockA{\textit{Department of Computer Science and Engineering} \\
\textit{Brac University}\\
Dhaka, Bangladesh \\
\{dibyo.fabian.dofadar, riyo.hayat.khan, md.sabbir.ahmed\}@g.bracu.ac.bd,\\
rafeedrahmansham2015@gmail.com}
}

\maketitle

\begin{abstract}
The rapid advancement of social networks and the convenience of internet availability have accelerated the rampant spread of false news and rumors on social media sites. Amid the COVID-19 epidemic, this misleading information has aggravated the situation by putting people’s mental and physical lives in danger. To limit the spread of such inaccuracies, identifying the fake news from online platforms could be the first and foremost step. In this research, the authors have conducted a comparative analysis by implementing five transformer-based models such as BERT, BERT without LSTM, ALBERT, RoBERTa, and a Hybrid of BERT \& ALBERT in order to detect the fraudulent news of COVID-19 from the internet. COVID-19 Fake News Dataset has been used for training and testing the models. Among all these models, the RoBERTa model has performed better than other models by obtaining an F1 score of 0.98 in both real and fake classes.
\end{abstract}


\section{Introduction}
One of the most difficult tasks in the field of Natural Language Processing is to detect misleading news, especially because of the vast expansion of social networks and easy accessibility of the internet. The massive online resources surely do include a lot of useful information, but at the same time, they consist of a lot of fraudulent news and rumors. Hence, it has become crucial to stop the spreading of misinformation, especially amid the COVID-19 crisis. \\
COVID-19 is undoubtedly one of the biggest catastrophes that have ever occurred in history. It is referred to as a global pandemic by the World Health Organization (WHO). While the whole world has been suffering within this chaos, people are forced to fight against “infodemic”. In this era of technological advancement, social media is playing a negative role by spreading fake news and rumors, resulting in creating panic among people. Since COVID-19 has involved the lives of people greatly, public attention has been focused on COVID-19-based information. Misusing this chance, many adversarial agents had intentionally spread bogus news for economic, political, or other purposes. A wave of myths, frauds, and misconceptions about the disease's etiology, effects, prevention, and treatment \cite{b11} had been widespread across the world of the internet.  Rumors like “spraying chlorine on your own body for preventing the coronavirus”, “drinking alcohol can kill the virus”, etc. \cite{b7} had been viral on various social media platforms. Believing this misleading information had put people in danger, in some cases, costing their lives. There had been a study by \cite{b7} in which it was estimated that almost 5800 people were hospitalized and 800 people died because of following the rumors on social media \cite{b10}. This widespread misinformation had not only damaged the people's lives physically but also mentally since it causes terror and anxiety \cite{b13}. \\
As a solution to this problem, one must detect misinformation from internet platforms. However, it is very complicated for humans to distinguish between false and true news since the process of evidence collection can be very tedious and also very time-consuming. Keeping all these in mind, a practical solution is to build models that could identify all the false information. In this research, the authors have implemented five transformer-based models such as BERT, BERT without LSTM, ALBERT, RoBERTa, and a Hybrid of BERT \& ALBERT for identifying fake news regarding COVID-19, and have performed a comparative analysis of those models. \\ 
Several pieces of research have been conducted to detect fraudulent news regarding COVID-19. In \cite{b9}, the authors had shown the use of Decision Tree, Logistic Regression, Gradient Boost, and Support Vector Machine (SVM) to detect fake news related to COVID-19. For this purpose, they prepared and annotated a dataset of more than ten thousand posts from social media and reports related to true and fake news about COVID-19. They had shown a binary classification between fake and real news. Among all the models that they used, the Support Vector Machine showed the highest value of 93.32\% in terms of F1 score when tested with the testing data. \\
In \cite{b3}, the automatic detection of whether a particular tweet related to COVID is fake or real was shown. The authors applied an ensemble method with pre-trained models for this objective. Also based upon username and link domains, they incorporated a unique heuristic algorithm. Genuine news items were obtained for the dataset from credible sources that supplied crucial COVID-19 information, whereas "fake" ones were gathered from tweets, posts, and articles that made false COVID-19 assertions. \\ 
A dataset of  Contraint@AAAI 2021 COVID-19 Fake news detection was used in \cite{b12}. The models that they have incorporated for fake news detection were CNN, LSTM, and BERT. \\
The authors of \cite{b2} have incorporated language models which are fine-tuned transformers with resilient loss functions and used influence calculation to remove harmful training instances. They evaluated the model on the COVID-19 misinformation test set (Tweets-19). They used Fake-News COVID-19 (FakeNews-19) which was the dataset made available for the collaborative effort of the CONSTRAINT 2021 workshop. There has been another research \cite{b6} in which the authors showed their best approach using BERT, ALBERT, and XLNET while detecting fake news related to COVID-19.

\section{Dataset}
\subsection{Data Description}
The dataset used in this research was referred to \cite{b9}. It contains 10,700 social media items and articles that have been manually annotated. This dataset was given two labels - 'real' and 'fake.' Out of the 10,700 instances, 6420 (60\%) social media posts and articles were separated for training and 2140 (20\%) social media posts and articles were chosen for testing and validation each. The total number of distinct words in the sample was 37,505 with 5141 words appearing in both actual and fraudulent news articles and posts. Furthermore, 52.34 percent of the samples were from credible news, while the remaining 47.66 percent were from fake news.

\subsection{Data Preprocessing and Exploratory Data Analysis}

At first, for convenience, the samples for ‘real’ labels were changed to 0 and the ‘fake’ labels were relabelled to 1. Then the major preprocessing steps were performed which included different text normalization and cleaning processes like stop words removal, link removal, lowercase transformation, contraction expansion, special characters removal, ignoration of ASCII values, translation table construction, etc. \\

After cleaning the data, it was checked whether there is any missing value in the dataset and it was found that there was no data loss in the text cleaning process. Figure 1 depicts the distribution of the number of words per post in train data. It can be seen that there are a very small number of posts whose length is large. So, to get the suitable sequence length, the number of words in the training data was quantiled. After a few tests, it was found that quantiling by 98\% and 99\% gives 52 and 56 length sequences respectively. This means if the maximum sequence length is set to 52, then no info will be lost in 98\% data and for 56, no info will be lost in 99\% data. For implementation purposes, as the models are computationally expensive, it was decided to take 56 as the maximum sequence length. \\

\begin{figure}[!h]
\begin{center}
\includegraphics[scale=0.5, width = 0.45\textwidth]{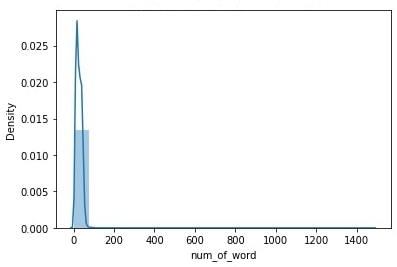} 

\caption{\centering Distribution of the number of words per post in train data}
\label{fig.1}
\end{center}
\end{figure}

Lastly, the distribution of labels can be visualized in Figure 2. It can be seen that data is nearly balanced across the training, test, and validation data. \\

\begin{figure*}[!h]
\centering
\begin{center}
\includegraphics[width = 0.75\textwidth]{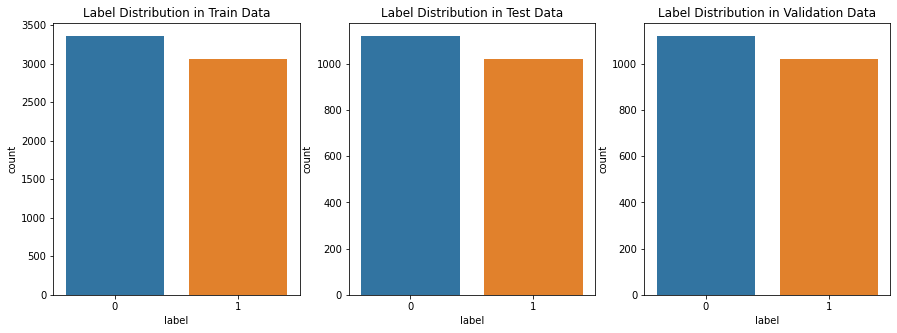} 

\caption{Distribution of labels in train, test and validation data}
\label{fig.2}
\end{center}
\end{figure*}

\begin{figure*}[!h]
\begin{center}
\includegraphics[scale=0.5, width = \textwidth, height = 12 cm]{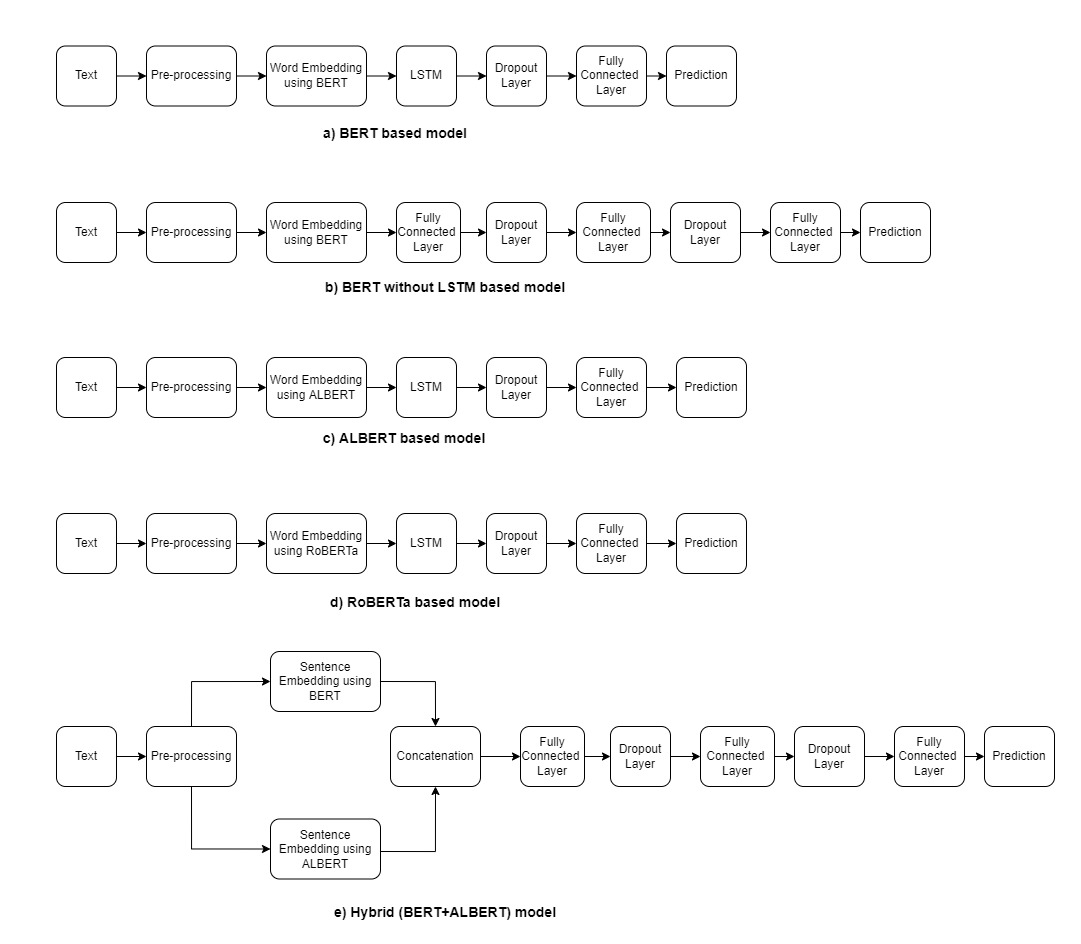} 

\caption{Block diagram of several implemented models}
\label{fig.3}
\end{center}
\end{figure*}

\section{Methodology} 
Transformer based approaches used in this research are BERT and other BERT inspired NLP architectures. The five different models implemented for this research are as follows:
\begin{itemize}
    \item \textbf{BERT based model:} BERT\cite{bert} is currently one of the most widely used attention-based linguistic interpretation models. BERT is built to pre-train deep bidirectional representations from unlabeled data by configuring both the left and right settings in all the layers. Consequently, with only one extra output layer, the pre-trained BERT model may be fine-tuned to develop cutting-edge models for a wide variety of tasks without requiring significant task-specific engineering changes. In the implementation, some extra preprocessing steps like tokenization and padding were applied before creating the model. The token inputs and mask inputs were jointly considered as the input for the model. BERT then generated the feature vectors of the text and this BERT output was passed to a 128-unit Long short-term memory (LSTM) \cite{lstm} layer followed by a dropout layer with a 20\% dropout rate. Finally, the model generated the final output using the attributes after a fully connected layer with the sigmoid activation function. Figure 3(a) shows the simple block diagram of the architecture of this model. 
    \item \textbf{BERT without LSTM based model:} In this model, after word embedding, the outputs of the BERT model were not followed by an LSTM layer. Instead, an extra 128-unit fully connected layer and a dropout layer with a 30\% dropout rate were connected after BERT output. Furthermore, another fully connected layer with 64 units was connected followed by another dropout layer with a 20\% dropout rate. Similar to the previous model, the final output was found after a fully connected layer with the sigmoid activation function. The simple architecture behind this model can be visualized in Figure 3(b).
    \item \textbf{ALBERT based model:} The ALBERT model\cite{albert} is a simplified BERT-based model for self-supervised language representation learning. It describes two parameter reduction strategies for lowering memory utilization and increasing BERT training speed. It divides the embedding matrix into two sub-matrices. It also uses repeated layers to separate between bunches. Architecturally it is almost similar to the first model. The implementation of this model for this research can be observed in Figure 3(c).

    \begin{table*}[hbt!]
\centering
\caption{\centering Precision, Recall, F1-score and Support of Real and Fake class for BERT, BERT without LSTM, ALBERT, RoBERTa and Hybrid (BERT+ALBERT) models}
\begin{tabular}{|l|llll|llll|}
\hline
                     & \multicolumn{4}{c|}{Fake Class}                                                                                      & \multicolumn{4}{c|}{Real Class}                                                                                      \\ \hline
Model Name           & \multicolumn{1}{l|}{Precision} & \multicolumn{1}{l|}{Recall} & \multicolumn{1}{l|}{F1 Score} & Support               & \multicolumn{1}{l|}{Precision} & \multicolumn{1}{l|}{Recall} & \multicolumn{1}{l|}{F1 Score} & Support               \\ \hline

BERT                 &\multicolumn{1}{l|}{0.87}      &\multicolumn{1}{l|}{0.88}   &\multicolumn{1}{l|}{0.88}       & 1020 &\multicolumn{1}{l|}{0.89}      &\multicolumn{1}{l|}{0.88}       &\multicolumn{1}{l|}{0.89}     &1120
\\ \cline{1-4} \cline{6-8}

BERT without LSTM    & \multicolumn{1}{l|}{0.77}      & \multicolumn{1}{l|}{0.71}   & \multicolumn{1}{l|}{0.74}     &                       & \multicolumn{1}{l|}{0.75}      & \multicolumn{1}{l|}{0.81}   & \multicolumn{1}{l|}{0.78}     &                       \\ \cline{1-4} \cline{6-8}
ALBERT               & \multicolumn{1}{l|}{0.97}      & \multicolumn{1}{l|}{0.87}   & \multicolumn{1}{l|}{0.92}     &                       & \multicolumn{1}{l|}{0.90}       & \multicolumn{1}{l|}{0.98}   & \multicolumn{1}{l|}{0.93}     &                       \\ \cline{1-4} \cline{6-8}
RoBERTa              & \multicolumn{1}{l|}{0.99}      & \multicolumn{1}{l|}{0.96}   & \multicolumn{1}{l|}{\textbf{0.98}}     &                       & \multicolumn{1}{l|}{0.97}      & \multicolumn{1}{l|}{0.99}   & \multicolumn{1}{l|}{\textbf{0.98}}     &                       \\ \cline{1-4} \cline{6-8}
Hybrid (BERT+ALBERT) & \multicolumn{1}{l|}{0.94}      & \multicolumn{1}{l|}{0.96}   & \multicolumn{1}{l|}{0.95}     &                       & \multicolumn{1}{l|}{0.96}      & \multicolumn{1}{l|}{0.94}   & \multicolumn{1}{l|}{0.95}     &                       \\ \hline
\end{tabular}
\end{table*}

\begin{table*}[hbt!]
\centering
\caption{\centering Comparison with Other Literature Approaches Applied on the Same Dataset}
\centering
\begin{tabular}{|l|l|l|l|l|}
\hline
Author's work & \multicolumn{1}{c|}{Methods}   & Precision     & Recall & F1-score      \\ \hline
\cite{b9}  & SVM                            & 0.9333        & 0.9332 & 0.9332        \\ \hline
\cite{b5}  & XLNet with Topic Distributions & 0.968         & 0.967  & 0.967         \\ \hline
\cite{b8}  & Cross-SEAN                     & 0.946         & 0.961  & 0.953         \\ \hline
\cite{b1}  & DistilBERT                     & 0.923         & 0.949  & 0.936         \\ \hline
\cite{b4}  & ERNIE 2.0                      & 0.976         & 0.976  & 0.976         \\ \hline
This work      & RoBERTa                        & \textbf{0.99} & 0.96   & \textbf{0.98} \\ \hline
\end{tabular}
\end{table*}
    
    \item \textbf{RoBERTa based model:} RoBERTa \cite{roberta} is also a BERT-inspired model in which the structure is constructed on BERT and then the model modifies key hyperparameters, removing the next-sentence pretraining purpose and preparing with significantly larger mini-batches and learning rates. This model has two major preferences over other transformer-based models. The expansive batches move forward with perplexity on masked dialect displaying objective and well as end-task accuracy. Also, huge batches are moreover less demanding to parallelize through distributed parallel training. In the implementation of this research, the structure of the model is similar to the first and third models in general. This model can be observed with a simple block diagram from Figure 3(d). 
    \item \textbf{Hybrid (BERT + ALBERT) model:} The last model was created by merging the outputs of the BERT and ALBERT models. In this model, no LSTM was used during implementation. Another major difference with other models was, that this model did not consider individual word’s feature vectors, but took feature vectors for the overall sentence. Thus a big feature vector was created by concatenating the embedding of BERT and ALBERT. This feature vector was fed into a 128-unit fully connected layer and a dropout layer with a 30\% dropout rate. Furthermore, another fully connected layer with 64 units was connected followed by another dropout layer with a 20\% dropout layer. Similar to the other models, the final output layer was created after a fully connected layer with the sigmoid activation function. Figure 3(e) represents the simple block diagram of this model architecture.
\end{itemize}
\section{Hyper parameter tuning}
Tensorflow API is used to build the models. For all the models, the hyper parameter was set under manual observation. A small portion of the data (40\% of training data and 30\% of validation data) was utilized for finding the best hyper parameters for the models. It was searched for an acceptable balance between bias and variance to get well suited hyper parameter. Adam optimizer \cite{adam} was applied on 32-sized batches and the learning rate was set at 0.00002 for every single model.
\section{Results}
As discussed earlier, a total of 5 models were used for the mentioned dataset namely BERT, BERT without LSTM, ALBERT, RoBERTa, and Hybrid (BERT+ALBERT).
From Table 1, it can be observed that the BERT model had achieved 87\% Precision, 88\% Recall, and F1 score of 88\% for the Fake class and 89\% Precision, 88\% Recall, and F1 score of 89\% for the Real class. For the BERT without LSTM the Precision, Recall, and F1 scores of the Fake class are 77\%, 71\%, and 74\% whereas for the Real class those are 75\%, 81\% and 78\% respectively. ALBERT shows an increase in all the metrics for both the Real and Fake classes. This model showed a Precision of 97\% along with 87\% Recall and 92\% F1 score for the Fake class and 90\% Precision, 98\% Recall, and 93\% F1 score for the Real class. A Hybrid model was also implemented for this work. This model was projected to outperform all other state-of-the-art models. Unfortunately, this model’s performance was not the best among the other models; instead, this model gave a decent result. For the Fake class, the Precision was 94\%, Recall was 96\% and the F1 score for this model was 95\%. For the Real class, the precision was 96\%, the recall was 94\% and the F1 score was 95\%. Among all models, the RoBERTa model was well suited for the dataset. It has the highest score for every metric. For the Real class, this model got 97\% Precision, 99\% Recall, and 98\% F1 scores. For the Fake class, the model achieved 99\% Precision, 96\% Recall, and 98\% F1 scores. The reason why RoBERTa performed better compared to other models was the progress of the expansive batches on masked dialect displaying objective and end-task accuracy. Moreover, these expansive batches are less challenging for parallelizing while using distributed parallel training. \\

\begin{figure}[!h]
\begin{center}
\includegraphics[scale=0.5, width = 0.48\textwidth]{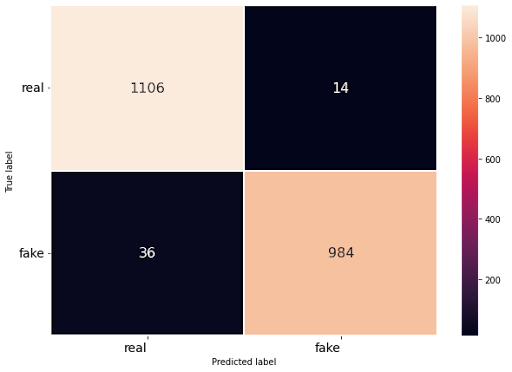} 

\caption{Confusion matrix of RoBERTa model}
\label{fig.4}
\end{center}
\end{figure}

In Figure 4, it can be seen that 2090 news was correctly classified for RoBERTa model. Here, 1106 news was real news and the rest 984 news were fake news were classified accurately. On the other end, 14 instances of real news were mislabeled as false news, while 36 instances of fake news were mislabeled as real news. \\
Table 2 compares the strategy to other authors in the literature that were used on the same dataset. According to the table, the suggested RoBERTa model outperforms existing techniques from similar literature implementations.

\section{Conclusion}
Widespread misinformation has a negative impact on people's lives, not just physically but also mentally, especially when it is linked to a global epidemic like COVID-19. As a result, detecting such misinformation becomes a crucial challenge in order to minimize the pandemonium as much as possible. Manually detecting fake news can be a time-consuming and daunting task. In this research, the authors suggested five different models for detecting both real and fake news in order to obtain more reliable, accurate, and simple detection of these fake news. In this research, RoBERTa has been the best fit for detecting both real and fake news whereas all other models have shown decent results. The authors have also proposed a hybrid model that incorporates BERT \& ALBERT models and demonstrated comparable outcomes to the state-of-the-art models. This paper is an attempt to use different approaches to recognize COVID-19 related false news to a significant extent. The future work of this research will include testing additional models as well as improving the hybrid model to detect COVID-19 related fake news and also handle similar cases better than existing methods.



\vspace{12pt}

\end{document}